\documentclass[conference]{IEEEtran}
\usepackage[colorlinks,urlcolor=blue,linkcolor=blue,citecolor=blue]{hyperref}

\usepackage{color,array}
\usepackage{multirow}
\usepackage{graphicx}

\ifCLASSINFOpdf
\else
\fi
\begin{document}
\title{One for All: An End-to-End Compact Solution for Hand Gesture Recognition}

\author{\IEEEauthorblockN{Monu Verma}
\IEEEauthorblockA{Computer Science and Engineering\\
Malaviya National Institute of Technology\\
Jaipur, India, 302017\\
Email: monuverma.cv@gmail.com}
\and
\IEEEauthorblockN{Ayushi Gupta}
\IEEEauthorblockA{ olx people, Bangalore,\\ Karnataka, 560034\\
Email: ayushigup26@gmail.com}
\and
\IEEEauthorblockN{Santosh K. Vipparthi}
\IEEEauthorblockA{Computer Science and Engineering\\
Malaviya National Institute of Technology\\
Jaipur, India, 302017\\
Email: skvipparthi@mnit.ac.in}}


%Telephone: (800) 555--1212, Fax: (888) 555--1212}


\maketitle

\begin{abstract}
The HGR is a quite challenging task as its performance is
influenced by various aspects such as illumination variations,
cluttered backgrounds, spontaneous capture, etc. The conventional CNN networks for HGR are following two stage pipeline to deal with the various challenges: complex signs, illumination variations, complex and cluttered backgrounds. The existing approaches needs expert expertise as well as auxiliary computation at stage 1 to remove the complexities from the input images. Therefore, in this paper, we proposes an novel end-to-end compact CNN framework: fine grained feature attentive network for hand gesture recognition (Fit-Hand) to solve the challenges as discussed above. The pipeline of the proposed architecture consists of two main units: FineFeat module and dilated convolutional (Conv) layer. The FineFeat module extracts fine grained feature maps by employing attention mechanism over multi-scale receptive fields. The attention mechanism is introduced to capture effective features by enlarging the average behaviour of multi-scale responses. Moreover, dilated convolution provides global features of hand gestures through a larger receptive field. In addition, integrated layer is also utilized to combine the features of FineFeat module and dilated layer which enhances the discriminability of the network by capturing complementary context information of hand postures. The effectiveness of Fit-Hand is evaluated by using subject dependent (SD) and subject independent (SI) validation setup over seven benchmark datasets: MUGD-I, MUGD-II, MUGD-III, MUGD-IV, MUGD-V, Finger Spelling and  OUHANDS, respectively. Furthermore, to investigate the deep insights of the proposed Fit-Hand framework, we performed ten ablation study  
\end{abstract}


% \fi
\IEEEpeerreviewmaketitle

\begin{figure}[!t]
\centering
\begin{center}
    	\includegraphics[width=\linewidth]{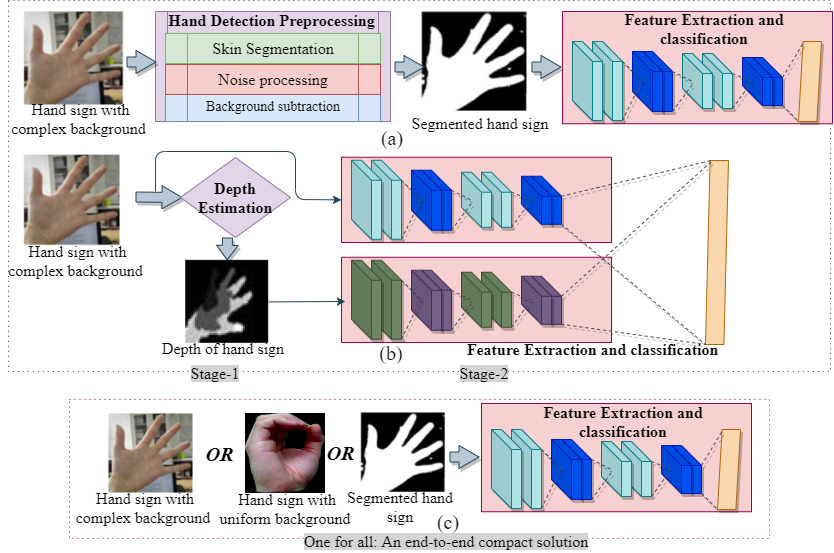}
	\caption{Architectural comparison between existing (a) \cite{chung2019efficient, oyedotun2017deep} -(b) \cite{yang2020ddanet,rakowski2018hand} two stage network and proposed one for all: an end-to-end solution for HGR. The visual representation implies the proposed frameworks efficacy to handle the all kind of challenges: complex signs, illumination variations, complex and cluttered backgrounds. While existing HGR frameworks aids extra computation to deal with the complex backgrounds.}
	\label{fig:Figure1.1}
\end{center}

\end{figure}
\begin{figure*}[!t]
\centering
\begin{center}
    	\includegraphics[width=\linewidth, height=1.4in]{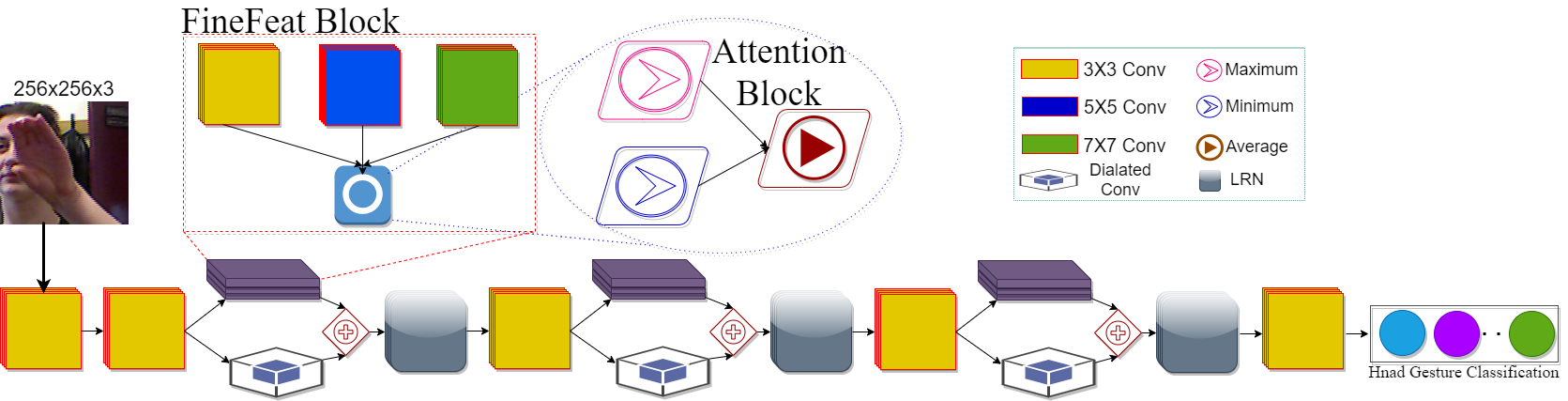}
	\caption{Architecture of the proposed Fit-Hand.}
	\label{fig:Figure1}
\end{center}

\end{figure*}
\section{Introduction}
 Hand gestures represent specific finger and hand movements that depict a particular message in non-verbal communication. Gestures also reinforce verbal communication by conveying human’s intentions in certain conversations. Hand gesture recognition is perceptual computing that allows machines to identify hand gestures and execute the relevant action. The current situation of coronavirus (COVID19) pandemic outbreak has caused sudden need of HGR in  various domains such as: consumer electronics market, transit sector, gaming, touch-less smartphones, defence, home automation, robotics, automated sign language translation etc. Thus, there is a need to amalgamate the HGR with AI to design and develop custom-made touch-less interface to carry out daily activities while maintaining physical distance. Thus, a robust HGR system is needed that can work efficiently on memory-limited devices for real-life applications. \par
 The HGR methods can be divided into two broad categories: sensor and vision-based techniques. Sensor based techniques \cite{sturman1994survey, santos2018dynamic} used gloves and other electronic devices to measure the joint angles position of the fingers, position of the hands to extract the features of hands.  Although, glove-based techniques have sufficient cues to identify hand gestures, but gloves with wires and sensors are too expensive and makes people uncomfortable to wear.  However, vision-based techniques can analyze hand gestures in a non-intrusive manner without any involvement of gloves and electromagnetic devices. Moreover, vision-based techniques are divided into two categories: 3D hand and appearance-based model. 3D hand-based models \cite{rehg1994visual,heap1996towards} represent hand structure by defining geometrical shapes of hands as joint angles of wrist, joints of fingers, space between fingers etc.  In appearance-based methods, texture features are extracted from visual appearance of hands. Appearance-based methods further can be split into two groups: pre-designed and learning based. Pre-designed based models encoded hand structure by imposing handcrafted feature descriptors.  Pre-designed feature descriptor \cite{pisharady2013attention, jasim2014sign} has got promising results in field of computer vision. However, these approaches fail to derive efficient feature in real time scenarios with variant challenges as noise, complex background conditions, low resolution and initial contour sequences.  Whereas, learning based approaches capture specific features of hand gestures by updating filter weights gradually.\par

Recently, convolutional neural networks (CNN) have shown tremendous performance in different computer vision fields like object detection, man pose estimation, anomaly detection, face verification, emotion recognition and many more. There are many deep CNN models postulated in literature such as Inception V3 \cite{szegedy2016rethinking}, ResNet \cite{he2016deep}, ResNet Inc \cite{szegedy2016inception}, DenseNet \cite{huang2017densely}, Mobilenet \cite{howard2017mobilenets}, Mobilenet V2 \cite{sandler2018mobilenetv2} and NasMobileNet \cite{zoph2018learning} etc. Even in HGR field, CNN based approaches \cite{zhan2019hand,adithya2020deep,mohanty2017deep, yahaya2019gesture} also have been shown impressive performance. These CNN based HGR frameworks are followed two-stage pipeline. Where, in first stage they have utilized handcrafted techniques to computes optical flow or heat maps. Whereas in second stage, CNN architectures are used for feature extraction and classification. Two-stage framework based approaches have gain impressive attention and successfully resolved the problem of multi-view, noise, low resolution etc. However, the performance of the these two stage models is limited by the pre-design techniques that are strongly dependent on prior expertise. To overcome the limitation of two-stage frameworks,  advanced CNN based approaches \cite{zhan2019hand,adithya2020deep,mohanty2017deep} have been introduced for hand gesture recognition without including any pre-designed feature extraction method. However, most of the advanced CNN models are  designed to solve specific problems like \cite{zhan2019hand} works for black and white hand posture images, \cite{adithya2020deep,mohanty2017deep} color images. Thus above mentioned work is not proving generic solution for all types of hand gesture images. Moreover, existing approaches were evaluated over subject dependent setup and have been gained high accuracy. However, they perform poorly when evaluated for unseen subjects' hand gestures (subject independent setup). The existing CNN networks also need huge computation with large parameters and incapable to work on handheld and portable devices.\par
Inspired with the above challenges, in this paper, one for all: an end-to-end compact network: fine grained feature attentive network for hand gesture recognition is introduced. The Fit-Hand model consist of two units: FineFeat module and dilated Conv layer to extract the effective fine features and abstract features, respectively. Furthermore, to learn the complementary information, resultant feature maps of FineFeat and dilated layer are combined by utilizing integrated layer. Complementary features allows network to learn both micro and high level features, which makes Fit-Hand a robust method to deal with black and white (segmented) as well as color full complex background hand gesture images. The proposed Fit-Hand also reduce the complexity of the HGR model by eliminating the need of hand segmentation. The visual demonstration of the comparison between state-of-the-art two-stage approaches and proposed one for all: FitHand framework is presented in Fig. \ref{fig:Figure1.1}. The main contribution of the proposed network is summarized as follows.

\begin{enumerate}
\item We proposed a light weighted end-to-end fine grained feature attentive network for hand gesture recognition as one solution for different challenges. 
\item A FineFeat module is proposed to extract abstract and detailed variation of the hand postures by utilizing both global and local receptive field information.
\item  A novel attention mechanism  is introduced to preserve effective edge information by utilizing averaging behavior of multi-scale receptive responses.
\item The dilated Conv layer is used to extract high-level feature of the hand-posture and improves the discriminability of the Fit-Hand.
\item Effectiveness of proposed Fit-Hand is validated on seven benchmark datasets: MUGD-I, MUGD-II, MUGD-III, MUGD-IV, MUGD-V, Finger Spelling (ASL), OUHANDS, with subject dependent and subject independent evaluation strategies.

\end{enumerate}

\section{Related work}
With recent advent of technologies CNN based approaches has gained good achievements in hand gesture recognition. Jose et al. \cite{flores2017application} designed two different CNN frameworks based on LeNET architecture \cite{lecun2015lenet} to extract the prominent features of hand gesture structures. Further, Oyebade et al. \cite{oyedotun2017deep} applied CNN network with auto-encoder to represents the features of hand gestures. Sérgio et al. \cite{chevtchenko2018convolutional} designed a feature fusion-based convolutional neural network (FFCNN) that incorporated auxiliary features extracted by gabor with CNN network. Moreover, Dadashzadeh et al. \cite{dadashzadeh2019hgr} proposed a two stage fusion network named as HGR Net. Where, in first stage, hand regions are detected by applying pixel-level semantic segmentation and second-stage network comprises two-stream CNN to determine the label of hand gesture. Furthermore, multi-task information sharing based approaches \cite{nie2018mutual,du2019crossinfonet} has been introduced for hand pose estimation. They extract the features of hands by decomposing them into sub-task through 2D and 3D- heat maps.
Mohanty et al. \cite{mohanty2017deep} proposed a deep learning model named as DeepGestures  for static hand gesture recognition. The DeepGestures model have been designed to handle the various challenges like variation in hand sizes, spatial location variations in the image and complex clutter background. Neethu et al. \cite{neethu2020efficient} introduced a CNN based hand gesture detection and recognition framework. Where, first they utilized mask images for hand region extraction and then segment fingers of images from the image though CNN. Further, the adaptive histogram equalization technique is used for image enhancement. Finally, the segmented fingers are fed to CNN model for hand gesture classification. Zhan et al. \cite{zhan2019hand} introduced a CNN model to solve the black and white hand gestures. Furthermore Islam  et al. \cite{islam2019static} utilized augmentation techniques and increase the hand gesture data sample to enhance the performance of CNN network. Adithya et al. \cite{adithya2020deep} introduce a CNN model for static hand gesture recognition without including any segmentation technique.

\section{Proposed Method}
Various researchers have exploit the learning capabilities of the pre-trained models \cite{cote2019deep, cote2017transfer} for HGR. Some of the existing approaches \cite{lin2014human,liao2018hand} have taken advantages of pre-designed feature descriptors and aid in CNN models to boost their performance. While, some of the CNN networks have been deigned to learn the features for specific hand postures. Furthermore, some other HGR approaches have achieve good results, but require huge computation cost. All above explained aspects limit the performance of HGR in practical scenarios.  This motivated us to design a generic and portable end-to-end CNN model for HGR which does not have dependency on neither pre-designed descriptors nor pre-trained weights. The detailed architecture of the proposed network is demonstrating in Fig. \ref{fig:Figure1}. 

Primarily network employ two consecutive Conv layers with $3\times3$ sized filters to extract variation patterns of hand poses. 
Let  $I\left(l,m\right)$ be an input image and $\eta_{S}^{u,v,d}\left(o \right)$  represents Conv function, where $S$ implies for stride, $d$ is depth, $u$ and $v$ represent the size of filter. Then response features $R_{f}$ of first two layers are calculated by Eq \ref{eq:1}.
\begin{equation}\label{eq:1}
    R_{f}=\eta_{2}^{3,3,32}\left \{\eta_{2}^{3,3,32}\left \{ I\left ( l,m \right ) \right \} \right \} 
\end{equation}

Further resultant feature maps are simultaneously forwarded to fine feature extraction (FineFeat) module and dilated Conv layer to preserve contextual information of hand postures.
\subsubsection{Fine Feature Extraction Module}
The aim of designing FineFeat module is to preserve fine grained edge information for discriminative feature representation of hand postures. The FineFeat module mainly comprises of three laterally connected multi-scale Conv of size $3\times3$, $5\times5$ and $7\times7$ with minimal parameters as show in Fig. \ref{fig:Figure1}. The multi-scale filters are liable to capture scale invariant features with multi-scale receptive fields. Further, attention block is employed  to fetch only effective edges and neglects others by establishing averaging concept over response multi-receptive fields. Response of FineFeat $\left(Im_{f}^{d}\right)$ module is calculated by using Eq. \ref{eq:2}.
\begin{equation}\label{eq:2}
    Im_{f}^{d}=\delta \left \{ \eta_{1}^{7\times7\times d}\left ( R_{f} \right ), \eta_{1}^{5\times5\times d}\left ( R_{f} \right ), \eta_{1}^{3\times3\times d}\left ( R_{f} \right )  \right \}
\end{equation}

where, $d$ represents depth of the Conv. Filters. attention block $\delta\left(o\right)$ is calculated by using Eq. $\left(\ref{eq:3}-\ref{eq:5}\right)$.
\begin{equation}\label{eq:3}
\delta \left( f_{1}, f_{2}, f_{3} \right)=\varphi \left( f_{1}, f_{2}, f_{3} \right)+min \left(\gamma\left( f_{1}, f_{2}, f_{3} \right)\right)  
\end{equation}
\begin{equation}\label{eq:4}
\gamma \left( f_{1}, f_{2}, f_{3} \right)= \left |\varphi \left( f_{1}, f_{2}, f_{3}\right)- \left( f_{1}, f_{2}, f_{3}\right) \right |   
\end{equation}
\begin{equation}\label{eq:5}
 \varphi \left( f_{1}, f_{2}, f_{3}\right)=\frac{1}{2} \left( max\left( f_{1}, f_{2}, f_{3}\right)+ min\left( f_{1}, f_{2}, f_{3}\right)\right)
\end{equation}

where, $ f_{1}, f_{2}, f_{3}$ are implies Conv layer holding multi-scale filters of size $3\times3$, $5\times5$ and $7\times7$ respectively.
\subsubsection{Dilated Convolution layer}
The dilated Conv layer \cite{yu2015multi} is embedded in FitHand network to extract global spatial features of hand gestures by refining inputs in high resolution. Dilated Conv layer allows to conserve more comprehensive context knowledge from input with reducing trainable parameters. Kernel size of dilated Conv is calculated by using Eq. \ref{eq:6}.
\begin{equation}\label{eq:6}
    R_{i}=i+\left ( i-1 \right )\left ( D-1 \right )
\end{equation}

Where, $i$ is the kernel size and $D$ represent the dilation rate. For Fit-Hand, we have used the 2 dilation. 
Moreover, Fit-Hand utilized the integrated layer \cite{he2016deep} to accumulate preserved feature maps of FineFeat module and dilated Conv layer. Integrated layer captures distinctive edge features and enhances robustness of Fit-Hand to define the disparities between different types of hand gestures problems. Final out-come of Fit-Hand can be computed by using Eq. $\left(\ref{eq:7}-\ref{eq:9}\right)$.
\begin{equation}\label{eq:7}
    H_{f}=FC\left [ \eta_{2}^{3\times3\times128}\left (  LRN\left \{ Im_{f}^{96}\left ( \chi_{1} \right )+Dil_{2}^{3\times3\times96} \left ( \chi_{1} \right )\right \} \right ) \right ]
\end{equation}
\begin{equation}\label{eq:8}
    \chi_{1}= LRN\left \{ Im_{f}^{64}\left ( \chi_{2} \right )+Dil_{2}^{3\times3\times64} \left ( \chi_{2} \right )\right \}
\end{equation}
\begin{equation}\label{eq:9}
    \chi_{2}= LRN\left \{ Im_{f}^{32}\left ( R_{f} \right )+Dil_{2}^{3\times3\times32} \left ( R_{f} \right )\right \}
\end{equation}

where, $Dil_{S}^{u\times v \times d}$  dilated convolution function, where $S$ implies for stride, $d$ is depth, $u$ and $v$ represent the size of filter. LRN and FC implies for local response normalization and fully connected layer.

%The effectiveness of FineFeat module and dilated Conv is depicted in Fig. \ref{fig:Figure2}, where the red line highlighted the influencive regions. From the Fig. \ref{fig:Figure2} it is clear that FineFeat module and Dilated Conv layer is capable to capture micro- level edges such as palm lines and high-level variations such as space finger lines, spaces between fingers, respectively. 
Since resultant responses are carrying different scale information,  local response normalization (LRN) and L2 normalization is used to normalize them. Therefore, these normalization techniques help to reduce the over-fitting and improve the prediction of the network.
\begin{table*}[t!]
\centering
\caption{Recognition Accuracy on MUGD, Finger Spelling, OUHANDS in SD and SI setups.\textit{Here, Fing. Spell, IncV3 and ResNet Inc, stands for finger spelling, inception V3 and resnet inception, respectively.}}
\label{tab:table2}
\vspace{-1em}
\begin{tabular}{|c|c|c|c|c|c|c|c|c|c|c|c|}
\hline
\multirow{3}{*}{\textbf{Method}} & \multicolumn{6}{c|}{\textbf{SD}}                                  & \multicolumn{5}{c|}{\textbf{SI}}                                                             \\ \cline{2-12} 
                        & \multicolumn{5}{c|}{\textbf{MUGD}} & \multirow{2}{*}{\textbf{Fing. Spell.}} & \multicolumn{3}{c|}{\textbf{MUGD}} & \multirow{2}{*}{\textbf{Fing. Spell.}} & \multirow{2}{*}{\textbf{OUHANDS}} \\ \cline{2-6} \cline{8-10}
                        & \textbf{I}  & \textbf{II}  & \textbf{III}  & \textbf{IV} & \textbf{V} &                               & \textbf{I}      & \textbf{II}      & \textbf{V}     &                               &  \\ \hline
{IncV3 \cite{szegedy2016rethinking} CVPR (2016)}     & {18.0} & { 23.5} & {11.5} & { 15.0} & { 12.0} & { 51.5}                                              & { 8.05} & { 8.80} & { 3.38} & { 25.5}   & { 34.8}                                           \\ \hline
{ ResNet50 \cite{he2016deep} CVPR (2016)}  & { 82.8} & { 89.2} & { 78.6} & { 83.2} & { 70.2} & { 95.0}                                              & { 66.2} & { 75.8} & { 37.7} & { 49.8} & { 63.4}  
\\ \hline
DeepGestures \cite{mohanty2017deep} CVIP (2016) &69.3& 83.0 &81.8&80.6&56.4&87.2&54.0&75.0&37.2&46.5& 46.3
\\ \hline
{ ResNetInc \cite{szegedy2016inception} AAAI (2017)} & { 45.0} & {50.6} & {39.2} & { 40.0} & { 36.2} & { 69.6}                                              & { 56.7} & { 33.6} & { 3.33} & { 22.8}  & {35.2}                                           \\ \hline
{ Dense121 \cite{huang2017densely} CVPR(2017)}  & { 82.0} & { 87.5} & { 72.0} & { 77.5} & { 72.1} & { 95.0}                                              & { 64.7} & { 69.2} & { 33.8} & { 55.7} & { 64.9}    \\ \hline
{ MobileNet \cite{howard2017mobilenets} (2017)}     & { 72.6} & { 81.4} & { 66.0} & { 69.8} & {68.4} & { 94.8}                                              & { 56.2} & {67.7} & {39.4} & { 51.0}   & { 59.0}                                           \\ \hline

DeepHand \cite{mohanty2017deep} Neu. Comp. (2017) &N/A &  N/A&N/A&N/A&N/A&91.33&N/A&N/A&N/A&N/A& N/A
\\\hline
{MobileV2 \cite{sandler2018mobilenetv2} CVPR (2018)} & { 50.6} & { 52.2} & { 38.2} & { 32.8} & { 42.6} & { 84.8}                                              & { 38.9} & { 37.2} & { 20.0} & { 45.4}   & {53.0}                                           \\ \hline
{ NASMob \cite{zoph2018learning} CVPR (2018)}    & { 21.5} & { 21.5} & { 18.5} & {19.5} & {15.5} & { 68.5}                                              & { 3.33} & { 3.60} & { 2.27} & { 37.8}   & { 37.9} 

\\ \hline

HandShape \cite{rakowski2018hand} ICCCV (2018) &N/A &  N/A&N/A&N/A&N/A&90.60&N/A&N/A&N/A&N/A& N/A
\\\hline
HandGes \cite{zhan2019hand} IRI (2019) &6.2 &6.4 &3.0&3.4&3.2&34.75&10.0&12.0&6.0&22.4& 23.0
\\

\hline
DeepConv \cite{adithya2020deep} PCS-Elsevier (2020) &71.2 &  80.4&74.8&86.4&63.4&\textit{96.0}&54.7&69.0&35.1&43.5& 56.0
\\
\hline
DDaNet \cite{yang2020ddanet} IEEE-Acc. (2020) &N/A &  N/A&N/A&N/A&N/A&94.10&N/A&N/A&N/A&N/A& N/A
\\
\hline
{\textbf{ Fit-Hand}}     & \textbf{ 85.2} & \textbf{ 91.6} & \textbf{ 98.6} & \textbf{ 98.8} & \textbf{ 72.2} & \textbf{ 95.8}                                              & \textbf{ 67.0} & \textbf{ 79.2} & \textbf{ 40.0} & \textbf{\ 58.8}    & \textbf{ 65.0}                                                                                       \\ \hline
\end{tabular}
\end{table*}

\subsection{Comparative Study with Existing Approaches}
The existing CNN networks: VGGNet \cite{simonyan2014very} and ResNet \cite{he2016deep} gain impressive results by using the sequential coupling behavior of Conv layers. However, in a deep dense network, linearly connected Conv layers may drop some salient features due to recurrence of cross-correlation, which has an important role to define a gesture class. Moreover, deep networks are failing to achieve good performance over smaller sized datasets \cite{schindler2016comparing}. To resolve this problem, we proposed a light-weighted end–to-end shallow network which is more appropriate in HGR systems. In addition, most of the challenging hand gesture datasets are captured with complex and cluttered backgrounds. Existing approaches needs hand segmentation to remove the complex background. Although, in literature various hand segmentation techniques like hand shape, skin, color segmentation etc. were proposed for hand segmentation. However, the same segmentation technique is not work with all types of backgrounds and limits the practical usability of the HGR. The proposed Fit-Hand utilized the integration layer to collect the complementary context features from FineFeat module and dilated Conv layer. FineFeat module provides fine grained features with the help of attention mechanism, which is able to capture effective edge information. While, dilated Conv layer generates global representation of hand gestures. Therefore, complementary features of FineFeat and dilated layer boosts the robustness of Fit-Hand to extract edges of hand postures and surpass the background information. Thus, the proposed Fit-Hand does not need hand segmentation. Also, Fit-Hand can easily learn the features from segmented or black and white hand gesture images. There, we conclude that Fit-Hand is a generic HGR framework that is capable to learn features in practical scenarios.
\begin{figure}[!t]
\centering
	\includegraphics[width=\linewidth,height=1.4in]{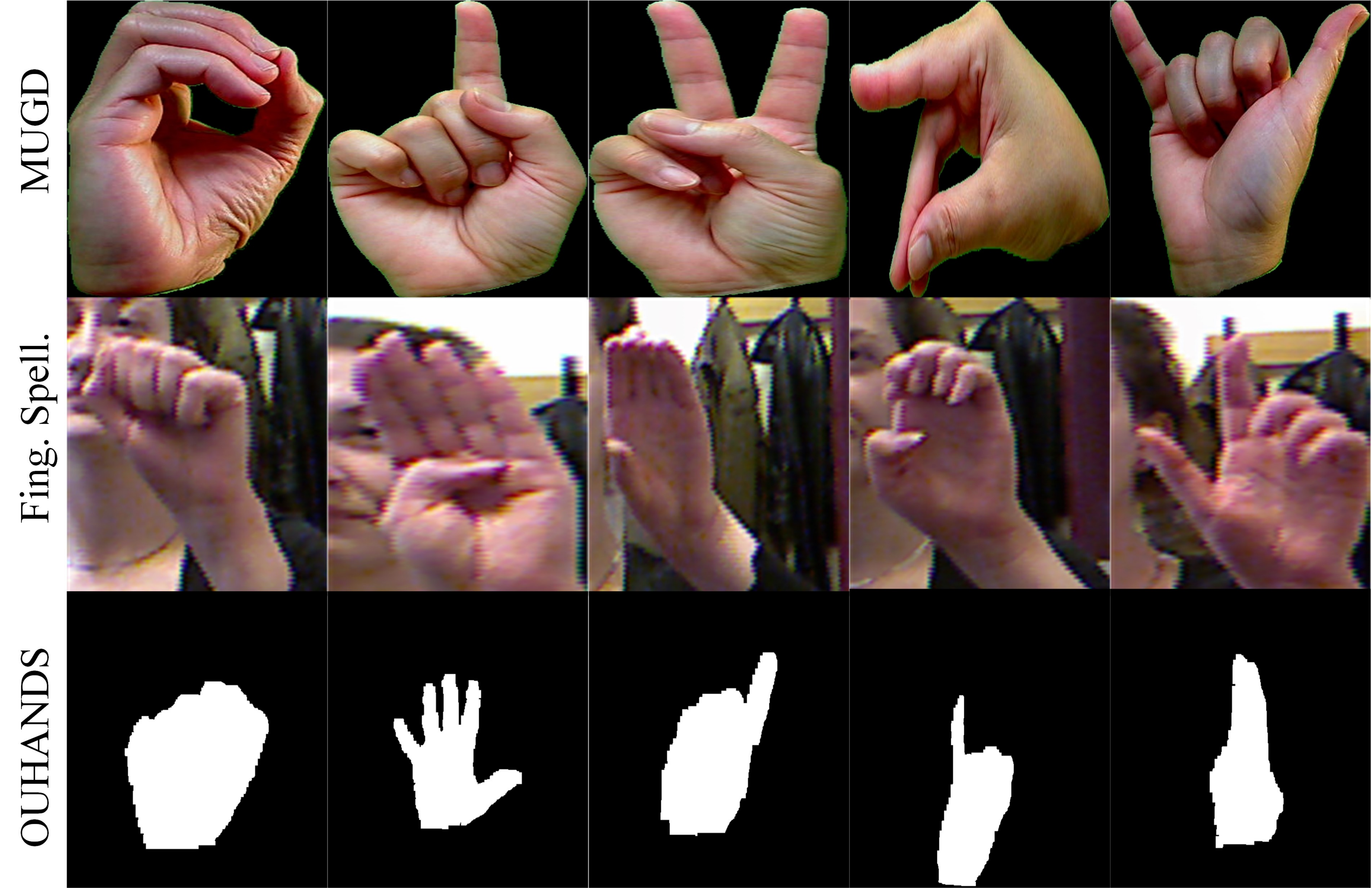}
	\caption{Sample images of different challenges as complex finger gestures, cluttered backgrounds, segmented gestures etc. in datasets: (a) MUGD, (b) Finger Spelling (ASL) and (c) OUHANDS, respectively.}
	\label{fig:Figure3}
\end{figure}

Fit-Hand incorporated a novel attention block, which has capability to extract only effective edges from the multi-scaled feature responses. Whereas, existing module inception layer \cite{szegedy2016inception}  simply concatenates previously extracted scale variant feature maps and let the neural network to learn relevant weights at the time of training. Thus, inception layer increases the complexity of network. Furthermore, Fit-Hand exploits the effectiveness of the dilated Conv layer and preserved the global context features of hand gestures.
\begin{figure*}[t!]
\centering
	\includegraphics[width=\linewidth, height=4.8in]{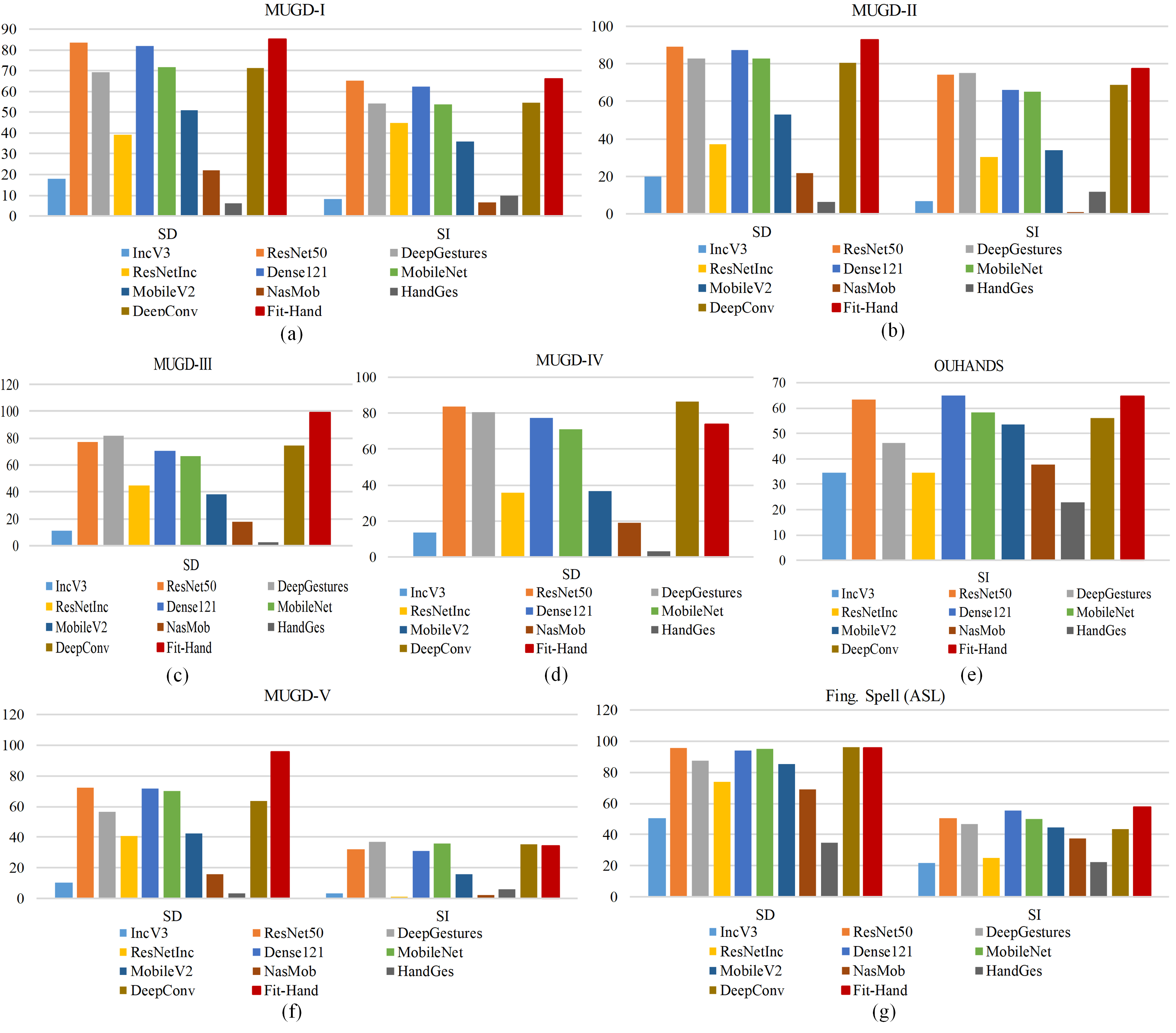}
	\caption{The graphical plot representing the comparative analysis between existing: IncV3, ResNet50, DeepGestures, ResNetInc, Dense121, MobileNet, MobileV2, NASMob, HandGes, DeepConv and proposed: Fit-Hand
in-terms of F1-Score over (a) MUGD-I, (b) MUGD-II, (c) MUGD-III, (d) MUGD-IV, (e) OUHANDS, (f) MUGD-V and (g) Finger Spelling (ASL) datasets for SD and SI experimental
settings, respectively. }
	\label{fig:Figure4}
\end{figure*}
Moreover, Fit-Hand embedded Conv layer with stride 2, to down-sample input size instead max pooling to preserve minute variation information. Max pooling executed max function to scale down the input size, which eliminates the minute edge features. Sometimes, small variations in a gesture may change the interpretation of its class, thus micro level edge variation representation is also playing a significant role to define a gesture. In literature some studies \cite{verma2019learnet, verma2019hinet} have validated that Conv with stride instead of pooling adds inter-feature dependencies and improves the learnability of neurons. 

\section{Experimental Setup and Analysis}
In this section, we examined the proposed network for
HGR on seven benchmark databasesdatasets: massey university gesture dataset part-I (MUGD-I), MUGD-II, MUGD-III, MUGD-IV, MUGD-V \cite{barczak2011new}, Finger Spelling (American Sign Language) \cite{pugeault2011spelling} and OUHANDs \cite{matilainen2016ouhands}. The quantitative and graphical results in terms of recognition accuracy and F1-Score is verified with the state-of-the-art methods for HGR. The qualitative results are demonstrated to visualise the effectiveness of Fit-Hand as compare to existing HGR approaches. Further, ten supplementary experiments are conducted for ablation study to validate the effectiveness of each module in the proposed method on OUHANDS. Furthermore, complexity analysis between proposed and state-of-the art models is represented to validate the potability of the Fit-Hand.

\subsection{Implementation details}
All experiments of Fit-Hand are conducted using Keras open-source deep-learning library with the Tensor flow in the backend.  The cross-entropy is used as the cost function and the SGD optimizer is used for optimization. The Fit-Hand is trained with learning rate 0.0001 for all experiments. The input image size has been fixed with $256\times256$ for training and inference of the model. The Nvidia GeForce RTX 2080 GPU
with Xeon processor, 16-core CPU, and 11 GB RAM under
Cuda 10.0. on Tensorflow-GPU 2.0.0 is used for the experiments. 

Moreover, to examine the effectiveness of the proposed Framework, we have compared our results with
other state-of-the-art approaches. The researchers have took up various dataset selection procedures
and experimental settings. Therefore, it is hard to make valid comparison between the various published results. To ensure fair comparison of HGR networks: DeepGestures, DeepConv and HandGes, we have implemented the all of them according to our experimental setups. In addition for general networks: Inception V3, ResNet50, ResNetInception, DenseNet, MobileNdet, MobileNetV2, NASMobileNet, we have fine tuned pre-trained weights with our experimental hyper-parameters over 10 epochs.
\begin{figure}[t!]
\centering
	\includegraphics[width=\linewidth, height=3.2in]{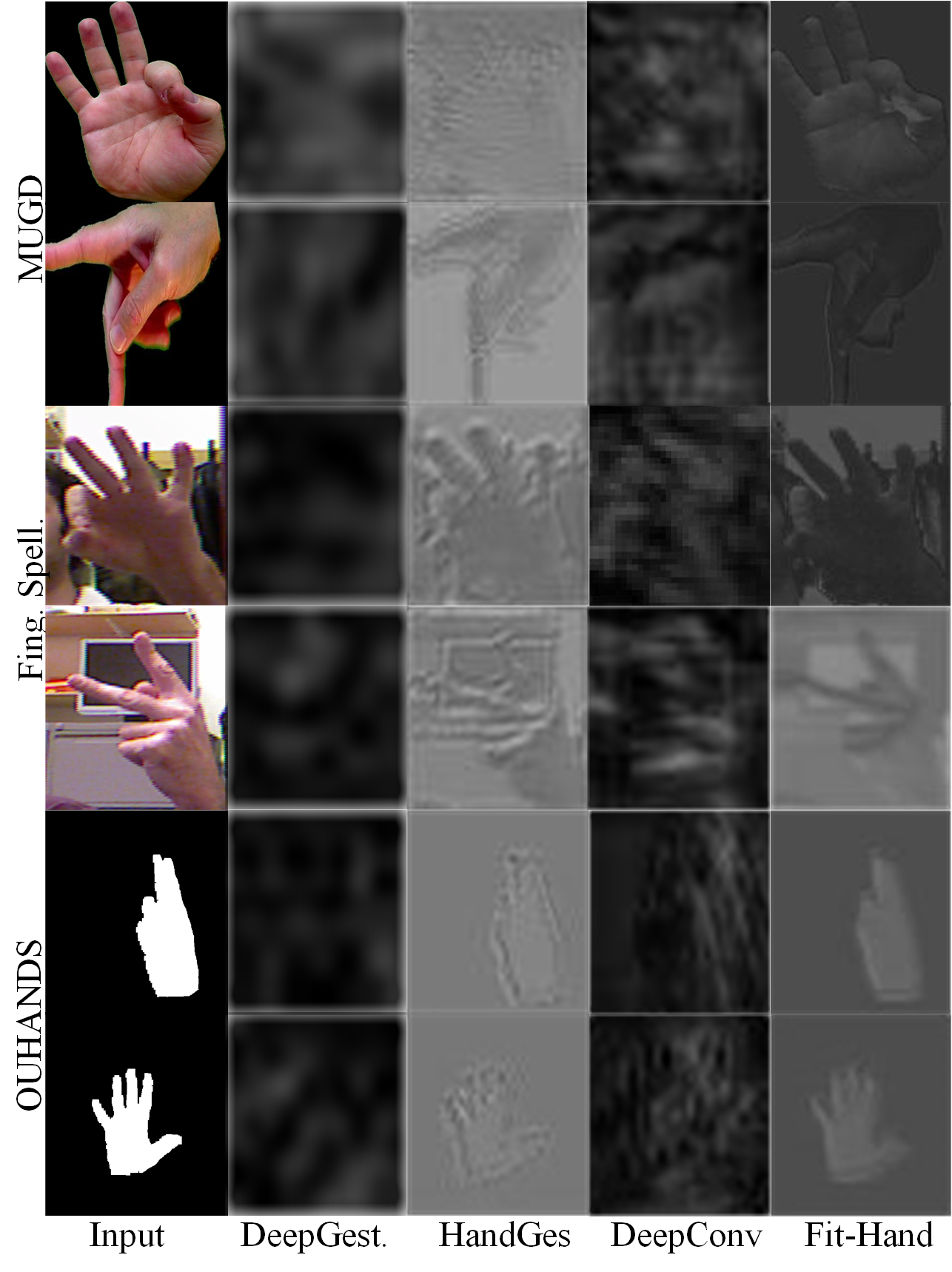}
	\caption{The qualitative comparison between the feature maps generated by state-of-the-art HGR network: DeepGestures, HandGes, DeepConv and proposed Fit-Hand, over six different gestures. }
	\label{fig:Figure5}
\end{figure}
As all versions of MUGD datatses are limited in number, and deep Convolutional neural networks require a large database to learn the most significant features, the datasets are augmented offline to enhance the generalization of the model and prevent over-fitting. The following transformations are applied for data augmentation: rotation in between [$-45 ^{\circ}$, $45^{\circ}$] with increment of $15^{\circ}$, horizontal flip and histogram equalization. Finally, one image instance is converted into 10 images.
 
\subsection{Experimental Setup}
In literature most of the researchers utilized N-fold cross validation scheme for evaluation. In N-fold cross validation, datasets are divided into random N folds, Where N-1 folds are used for training purpose and one fold is used for inference. The same procedure is followed for each fold and average of all folds are considered as final results. However, N-fold cross validation strategy is a subject dependent evaluation due to random division of folds and not ensure the performance of models for unseen data. Therefore, these approaches have been gained high accuracy for seen data samples. However, they perform poorly when evaluated for unseen subjects' hand gestures (subject independent setup) and not suitable for real time data validation. Thus, for fair  performance of the Fit-Hand we have adopted two validation schemes: subject- dependent and -independent. In subject dependent (SD), datasets are randomly partitioned into 80:20 ration such that 80$\%$ dataset is processed for training and 20$\%$ for testing set. While, in subject independent (SI), three subjects’ hand gestures are used in training and remaining hand gestures are used for inference for MUGD-I, MUGD-II and Finger Spelling datasets. For MUGD-V hand gesture, one subject is included in training set and other one used for testing purpose. The data division for SI is done purely in mutually exclusive manner. Moreover, OUHANDS dataset contains two parts: training and testing set with 2000 and 1000 images, respectively. 

\subsection{Quantitative Analysis}
This section demonstrates the effectiveness of proposed network over all datasets: MUGD, Finger Spelling and OUHANDS, in terms of recognition accuracy over two experimental setups: SD and SI respectively. Comparative analysis of existing and Fit-Hand is tabulated in Table \ref{tab:table2}. Specifically, Fit-Hand achieved 15.9\%, 8.6\%, 16.8\%, 18.2\%, 15.8\% and 14\%, 11.2\%, 23.8\%, 12.4\%, 8.8\% more accuracy as compared to DeepGestures and DeepConv HGR models over MUGD dataset for part I-V in SD setup, respectively. 
 Similarly, in SI setup, Fit-Hand gained 13\%, 4.2\%, 2.8\%, 12.3\%, 18.7\% and 12.3\%, 10.2\%, 4.9\%, 15.3\%, 9\% more accuracy as compare to DeepGestures and DeepConv models for MUGD-I, MUGD-II, MUGD-V, Finger Spelling and OUHANDS datasets. Moreover, performance of Fit-Hand in terms of F1-Score are graphically demonstrated in Fig. \ref{fig:Figure4}. From the Table \ref{tab:table2} and Fig. \ref{fig:Figure4} results it is clear that all HGR methods: proposed as well as state-of-the-art generates high results in subject dependent setup as compare to subject independent. Moreover, the proposed FitHand outperformed the two-stage networks; DeepHand, HandShape and DDaNet with 4.45\%, 5.2\% and 1.7\% high accuracy over ASL finger spelling dataset. From the results, it is validated that proposed framework is robust to all kind of challenges presents in the HGR, which reflect the efficacy of the model to real-life applications. Also from the results, it is proven that subject independent validation strategy is more significant as compare to subject dependent to examined the performance of any CNN model. In addition some methods like Inception V3, NasMobile and HandGes are under-fitted and not suitable for small size datasets. 

\begin{figure}[t!]
\centering
	\includegraphics[width=\linewidth, height=3.6in]{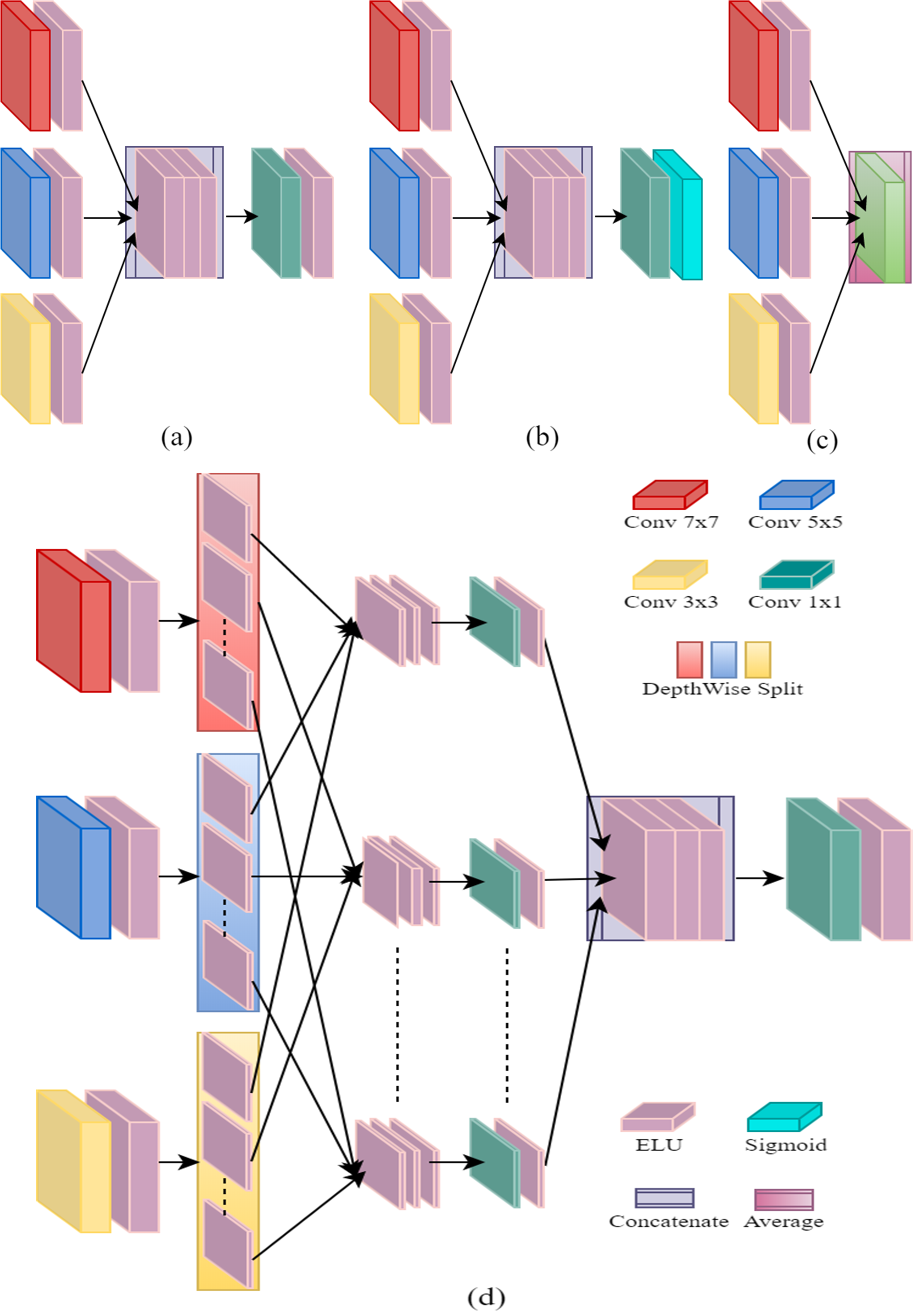}
	\caption{Different structures of FineFeat module for ablation study a) FineFeat with concatenation (FineFeat$\_$Cat), b) FineFeat with concatenation and Sigmoid (fineFeat$\_$CatSig), c) FineFeat with average (FineFeat$\_$Avg) and d) FineFeat with deep median (FineFeat$\_$DpMed).}
	\label{fig:Figure6}
\end{figure}

\subsection{Qualitative Analysis}
This section elaborates the effectiveness of Fit-Hand through the visual representation of neurons. Fig. \ref{fig:Figure5} depicted response maps of two different gestures captured at first Conv layer of ResNet-50, ResNet-Inc, MobileNet, MobileNet V2 and Fit-Hand. To represent the significance of all response maps, we have calculated mean response for each network. From the figure it is clear that Fit-Hand extracts more fine edge variations and highlighted regions like figure lines, palm lines, thumb articulates etc, which plays a significant role to define distinctiveness between hand postures. From above, it is clear that Fit-Hand has capability to preserve prominent features of hand gestures. Therefore, we can conclude that Fit-Hand has preserved more relevant feature responses to outperform the existing CNN based networks ResNet\-50, Res-Net-Inc, MobileNet and MobileNet V2 for different hand postures.
\subsection{Ablation Study}
In order to investigate the deep insights of Fit-HandeNet, 
we have conducted ten more ablation experiments for detail study as represented in Table \ref{tab:table4} over OUHANDS Dataset. This section fully explores the contribution of each module (FineFeat module, dilation layer, L2-Normalization, loss function and attention block) of the net-work in terms of performance and network complexity.
\begin{table}[t!]
\centering
\caption{Ablation results in terms of accuracy and complexity.\textit{Here, Param, Acc, M, S and MB stands for parameters, accuracy, millions, seconds and megabytes.}}
\label{tab:table4}
\vspace{-1em}
\begin{tabular}{|c|c|c|c|c|}
\hline
\textbf{Method}  & \textbf{Acc.}& \textbf{{\begin{tabular}[c]{@{}c@{}}\#Param \\ (M)\end{tabular}}}               & \textbf{{\begin{tabular}[c]{@{}c@{}}\#Mem. \\ (MB)\end{tabular}}}& \textbf{\#Time (S)}                  \\ \hline
{Fit-Hand\_WImp}    & {61.7} & {0.5} & {3.3}  & {1.08} \\ \hline
{Fit-Hand\_WDil}    & {60.9} & {1.4} & {11.9} & {2.32} \\ \hline
{Fit-Hand\_WL}      & {47.3} & {1.8} & {12.9} & {3.26} \\ \hline
{Fit-Hand\_2Stack}  & {51.7} & {1.0} & {6.5}  & {.82} \\ \hline
{Fit-Hand\_4Stack}  & {62.0} & {3.4} & {27.1} & {5.07} \\ \hline
{Fit-Hand\_Kul}     & {61.7} & {1.8} & {12.9} & {2.75} \\ \hline
{FineFeat\_Cat}    & {57.9} & {1.6} & {13.3} & {2.58} \\ \hline
{FineFeat\_CatSig} & {60.1} & {1.6} & {13.3} & {2.66} \\ \hline
{FineFeat\_Avg}    & {58.6} & {1.5} & {12.9} & {2.63} \\ \hline
{FineFeat\_DpMed}  & {58.3} & {1.6} & {16.0} & {30.8} \\ \hline
\textbf{ Fit-Hand}          & \textbf{ 65.0} & \textbf{ 1.8} & \textbf{12.9} & \textbf{ 4.00} \\ \hline
\end{tabular}
\end{table}
Specifically, to validate the importance of FineFeat module and dilated layer in Fit-Hand, we have evaluated results for Fit-Hand without FineFeat module (Fit-Hand$\_$WImp) and Fit-Hand without dilated layer (Fit-Hand$\_$WDil). From the Table \ref{tab:table4}, it is clear that both FineFeat module and dilated layer play a significant role in Fit-Hand and improve the performance of the network. To analyze the role of L2 normalization, results are evaluated by dropping L2 normalization (Fit-Hand$\_$WL). Evaluated results in Table \ref{tab:table4}, validated the effect of L2 normalization with high performance.  

To investigate that how three stacks of FineFeat modules help to learn the adequate information in Fit-Hand, we have performed two supplementary experiments with 2 FineFeat module (2 stacked) and Fit-Hand with 4 FineFeat module (4 Stacked). From the results tabulated in Table \ref{tab:table4}, it is concluded that, proposed Fit-Hand outperforms other dept combinations of FineFeat module. Moreover, to validate the effectiveness of cross-entropy loss function, we have computed results for Fit-Hand by replacing cross-entropy by Kullback Leibler Divergence Loss (Fit-Hand$\_$Kul). Computed results confirmed that cross-entropy loss function is most suitable for hand gestures classification in Fit-Hand.  

Furthermore, to examine the performance of attention block, we have implemented four different FineFeat modules by replacing pivot with a) concatenation (FineFeat$\_$Cat), b) concatenation with sigmoid (FineFeat$\_$CatSig) , c) average (FineFeat$\_$Avg) and d) deep median (FineFeat$\_$DpMed), as shown in Fig. \ref{fig:Figure6}. From Table \ref{tab:table4}, it is evident that proposed FineFeat module with attention block outperforms all other combinations of the FineFeat module.

\begin{table}[t!]
\centering
\caption{Complexity analysis comparison between existing and proposed Fit-Hand network.\textit{Here, M, K, MB, KB and S represents millions, thousands, megabytes, kilobytes and seconds, respectively.}}
\label{tab:table5}
\vspace{-1em}
\begin{tabular}{|c|c|c|c|}
\hline
\textbf{Method}                 &\textbf{\#Param  }             & \textbf{\#Mem.} & \textbf{\#Time (S)}                    \\ \hline
IncV3 \cite{szegedy2016rethinking}      & {22M}  & {179.3MB} & {17.64} \\ \hline
ResNet50 \cite{he2016deep}  & {31M}  & {208.4MB} & {18.17} \\ \hline
DeepGestures \cite{mohanty2017deep} & \textit{10K}  & \textit{128KB} & \textit{0.48} \\ \hline
{ResNetInc \cite{szegedy2016inception}} & {4M}   & {40.5MB}  & {60.93} \\ \hline
 Dense121 \cite{huang2017densely} & {7.5M} & {61.3MB}  & {24.34} \\ \hline
Mobile \cite{howard2017mobilenets}  & {5M}   & {36.5MB}  & {8.26}  \\ \hline
{MobileV2 \cite{sandler2018mobilenetv2}} & {3.5M} & {24.4MB}  & {15.26} \\ \hline
{NASMob \cite{zoph2018learning}}    & {4.7M} & {41.7MB}  & {45.65} \\ \hline
HandGes \cite{zhan2019hand} & {16K}  & {252KB} & {0.67} \\ \hline
DeepConv \cite{adithya2020deep} & {1M}  & {1.32MB} & {1.00} \\ \hline
\textbf{Fit-Hand}            & \textbf{1.8M} & \textbf{12.9MB}  & \textbf{4.00}  \\ \hline
\end{tabular}
\end{table}
\subsection{Complexity Analysis}
This section provides a comparative analysis of the computational complexity between the existing and proposed network. The total number of parameters, memory space and testing time involved in each network are tabulated in Table \ref{tab:table5}. The proposed Fit-Hand has very lesser number of parameters 1.8M as compared to other state-of-the-art models like: Inception V3: 22M, ResNet\-50: 31M, ResNet\-Inception: 4M, DenseNet 121: 7.5M, MobileNet: 5M, MobileNet V2: 3.5M and NASNet Mobile: 4.7M. Moreover, Fit-Hand trainable model captures smaller memory storage as compared to others. Fit-Hand query response time is also very less as compare to existing approaches. However, from the Table \ref{tab:table5}, it is also observable that complexity of some existing HGR models  [\cite{adithya2020deep}\cite{mohanty2017deep}\cite{zhan2019hand}] is less as compared to proposed FitHand. However, it is clear from Table \ref{tab:table2} that, these approaches are not providing generic solution as they failed to maintain good performance for all datasets. Therefore, on the basis of experimental and computational complexity results, we conclude that the proposed FitHand framework is a portable and generic solution for the different hand gestures. 
\section{Conclusion}
We proposed an one for all: end-to-end compact solution named as Fit-Hand: fine grained feature attentive network for HGR, which is responsible to identify distinct classes of hand gestures. Fit-Hand contains two main blocks: FineFeat module and dilated Conv layer.  FineFeat module conserves features of minute as well as major edge variation regions and further employ a attention block that has ability to fetch only pertinent features. Similarly, dilated layer is incorporated to capture global features. Further, Integrated layer added feature map of both blocks and enhance the learnability of Fit-Hand. Cohesively both layers allow Fit-Hand to learn the imperative features of hand postures and define disparities between them. Furthermore, variants of Fit-Hand were evaluated to verify the effectiveness of proposed Fit-Hand. 

%\bibliographystyle{IEEEtran}

\bibliographystyle{IEEEtran}
\bibliography{egbib}

\begin{thebibliography}{10}
\providecommand{\url}[1]{#1}
\csname url@samestyle\endcsname
\providecommand{\newblock}{\relax}
\providecommand{\bibinfo}[2]{#2}
\providecommand{\BIBentrySTDinterwordspacing}{\spaceskip=0pt\relax}
\providecommand{\BIBentryALTinterwordstretchfactor}{4}
\providecommand{\BIBentryALTinterwordspacing}{\spaceskip=\fontdimen2\font plus
\BIBentryALTinterwordstretchfactor\fontdimen3\font minus
  \fontdimen4\font\relax}
\providecommand{\BIBforeignlanguage}[2]{{%
\expandafter\ifx\csname l@#1\endcsname\relax
\typeout{** WARNING: IEEEtran.bst: No hyphenation pattern has been}%
\typeout{** loaded for the language `#1'. Using the pattern for}%
\typeout{** the default language instead.}%
\else
\language=\csname l@#1\endcsname
\fi
#2}}
\providecommand{\BIBdecl}{\relax}
\BIBdecl

\bibitem{chung2019efficient}
H.-Y. Chung, Y.-L. Chung, and W.-F. Tsai, ``An efficient hand gesture
  recognition system based on deep cnn,'' in \emph{2019 IEEE International
  Conference on Industrial Technology (ICIT)}.\hskip 1em plus 0.5em minus
  0.4em\relax IEEE, 2019, pp. 853--858.

\bibitem{oyedotun2017deep}
O.~K. Oyedotun and A.~Khashman, ``Deep learning in vision-based static hand
  gesture recognition,'' \emph{Neural Computing and Applications}, vol.~28,
  no.~12, pp. 3941--3951, 2017.

\bibitem{yang2020ddanet}
S.-H. Yang, W.-R. Chen, W.-J. Huang, and Y.-P. Chen, ``Ddanet: Dual-path
  depth-aware attention network for fingerspelling recognition using rgb-d
  images,'' \emph{IEEE Access}, 2020.

\bibitem{rakowski2018hand}
A.~Rakowski and L.~Wandzik, ``Hand shape recognition using very deep
  convolutional neural networks,'' in \emph{Proceedings of the 2018
  International Conference on Control and Computer Vision}, 2018, pp. 8--12.

\bibitem{sturman1994survey}
D.~J. Sturman and D.~Zeltzer, ``A survey of glove-based input,'' \emph{IEEE
  Computer graphics and Applications}, vol.~14, no.~1, pp. 30--39, 1994.

\bibitem{santos2018dynamic}
L.~Santos, N.~Carbonaro, A.~Tognetti, J.~L. Gonz{\'a}lez, E.~De~la Fuente,
  J.~C. Fraile, and J.~P{\'e}rez-Turiel, ``Dynamic gesture recognition using a
  smart glove in hand-assisted laparoscopic surgery,'' \emph{Technologies},
  vol.~6, no.~1, p.~8, 2018.

\bibitem{rehg1994visual}
J.~M. Rehg and T.~Kanade, ``Visual tracking of high dof articulated structures:
  an application to human hand tracking,'' in \emph{European conference on
  computer vision}.\hskip 1em plus 0.5em minus 0.4em\relax Springer, 1994, pp.
  35--46.

\bibitem{heap1996towards}
T.~Heap and D.~Hogg, ``Towards 3d hand tracking using a deformable model,'' in
  \emph{Proceedings of the Second International Conference on Automatic Face
  and Gesture Recognition}.\hskip 1em plus 0.5em minus 0.4em\relax Ieee, 1996,
  pp. 140--145.

\bibitem{pisharady2013attention}
P.~K. Pisharady, P.~Vadakkepat, and A.~P. Loh, ``Attention based detection and
  recognition of hand postures against complex backgrounds,''
  \emph{International Journal of Computer Vision}, vol. 101, no.~3, pp.
  403--419, 2013.

\bibitem{jasim2014sign}
M.~Jasim and M.~Hasanuzzaman, ``Sign language interpretation using linear
  discriminant analysis and local binary patterns,'' in \emph{2014
  International Conference on Informatics, Electronics \& Vision
  (ICIEV)}.\hskip 1em plus 0.5em minus 0.4em\relax IEEE, 2014, pp. 1--5.

\bibitem{szegedy2016rethinking}
C.~Szegedy, V.~Vanhoucke, S.~Ioffe, J.~Shlens, and Z.~Wojna, ``Rethinking the
  inception architecture for computer vision,'' in \emph{Proceedings of the
  IEEE conference on computer vision and pattern recognition}, 2016, pp.
  2818--2826.

\bibitem{he2016deep}
K.~He, X.~Zhang, S.~Ren, and J.~Sun, ``Deep residual learning for image
  recognition,'' in \emph{Proceedings of the IEEE conference on computer vision
  and pattern recognition}, 2016, pp. 770--778.

\bibitem{szegedy2016inception}
C.~Szegedy, S.~Ioffe, V.~Vanhoucke, and A.~Alemi, ``Inception-v4,
  inception-resnet and the impact of residual connections on learning,''
  \emph{arXiv preprint arXiv:1602.07261}, 2016.

\bibitem{huang2017densely}
G.~Huang, Z.~Liu, L.~Van Der~Maaten, and K.~Q. Weinberger, ``Densely connected
  convolutional networks,'' in \emph{Proceedings of the IEEE conference on
  computer vision and pattern recognition}, 2017, pp. 4700--4708.

\bibitem{howard2017mobilenets}
A.~G. Howard, M.~Zhu, B.~Chen, D.~Kalenichenko, W.~Wang, T.~Weyand,
  M.~Andreetto, and H.~Adam, ``Mobilenets: Efficient convolutional neural
  networks for mobile vision applications,'' \emph{arXiv preprint
  arXiv:1704.04861}, 2017.

\bibitem{sandler2018mobilenetv2}
M.~Sandler, A.~Howard, M.~Zhu, A.~Zhmoginov, and L.-C. Chen, ``Mobilenetv2:
  Inverted residuals and linear bottlenecks,'' in \emph{Proceedings of the IEEE
  conference on computer vision and pattern recognition}, 2018, pp. 4510--4520.

\bibitem{zoph2018learning}
B.~Zoph, V.~Vasudevan, J.~Shlens, and Q.~V. Le, ``Learning transferable
  architectures for scalable image recognition,'' in \emph{Proceedings of the
  IEEE conference on computer vision and pattern recognition}, 2018, pp.
  8697--8710.

\bibitem{zhan2019hand}
F.~Zhan, ``Hand gesture recognition with convolution neural networks,'' in
  \emph{2019 IEEE 20th International Conference on Information Reuse and
  Integration for Data Science (IRI)}.\hskip 1em plus 0.5em minus 0.4em\relax
  IEEE, 2019, pp. 295--298.

\bibitem{adithya2020deep}
V.~Adithya and R.~Rajesh, ``A deep convolutional neural network approach for
  static hand gesture recognition,'' \emph{Procedia Computer Science}, vol.
  171, pp. 2353--2361, 2020.

\bibitem{mohanty2017deep}
A.~Mohanty, S.~S. Rambhatla, and R.~R. Sahay, ``Deep gesture: static hand
  gesture recognition using cnn,'' in \emph{Proceedings of International
  Conference on Computer Vision and Image Processing}.\hskip 1em plus 0.5em
  minus 0.4em\relax Springer, 2017, pp. 449--461.

\bibitem{yahaya2019gesture}
S.~W. Yahaya, A.~Lotfi, M.~Mahmud, P.~Machado, and N.~Kubota, ``Gesture
  recognition intermediary robot for abnormality detection in human
  activities,'' in \emph{2019 IEEE Symposium Series on Computational
  Intelligence (SSCI)}.\hskip 1em plus 0.5em minus 0.4em\relax IEEE, 2019, pp.
  1415--1421.

\bibitem{flores2017application}
C.~J.~L. Flores, A.~G. Cutipa, and R.~L. Enciso, ``Application of convolutional
  neural networks for static hand gestures recognition under different
  invariant features,'' in \emph{2017 IEEE XXIV International Conference on
  Electronics, Electrical Engineering and Computing (INTERCON)}.\hskip 1em plus
  0.5em minus 0.4em\relax IEEE, 2017, pp. 1--4.

\bibitem{lecun2015lenet}
Y.~LeCun \emph{et~al.}, ``Lenet-5, convolutional neural networks,'' \emph{URL:
  http://yann. lecun. com/exdb/lenet}, vol.~20, no.~5, p.~14, 2015.

\bibitem{chevtchenko2018convolutional}
S.~F. Chevtchenko, R.~F. Vale, V.~Macario, and F.~R. Cordeiro, ``A
  convolutional neural network with feature fusion for real-time hand posture
  recognition,'' \emph{Applied Soft Computing}, vol.~73, pp. 748--766, 2018.

\bibitem{dadashzadeh2019hgr}
A.~Dadashzadeh, A.~T. Targhi, M.~Tahmasbi, and M.~Mirmehdi, ``Hgr-net: a fusion
  network for hand gesture segmentation and recognition,'' \emph{IET Computer
  Vision}, vol.~13, no.~8, pp. 700--707, 2019.

\bibitem{nie2018mutual}
X.~Nie, J.~Feng, and S.~Yan, ``Mutual learning to adapt for joint human parsing
  and pose estimation,'' in \emph{Proceedings of the European Conference on
  Computer Vision (ECCV)}, 2018, pp. 502--517.

\bibitem{du2019crossinfonet}
K.~Du, X.~Lin, Y.~Sun, and X.~Ma, ``Crossinfonet: Multi-task information
  sharing based hand pose estimation,'' in \emph{Proceedings of the IEEE
  Conference on Computer Vision and Pattern Recognition}, 2019, pp. 9896--9905.

\bibitem{neethu2020efficient}
P.~Neethu, R.~Suguna, and D.~Sathish, ``An efficient method for human hand
  gesture detection and recognition using deep learning convolutional neural
  networks,'' \emph{Soft Computing}, pp. 1--10, 2020.

\bibitem{islam2019static}
M.~Z. Islam, M.~S. Hossain, R.~ul~Islam, and K.~Andersson, ``Static hand
  gesture recognition using convolutional neural network with data
  augmentation,'' in \emph{2019 Joint 8th International Conference on
  Informatics, Electronics \& Vision (ICIEV) and 2019 3rd International
  Conference on Imaging, Vision \& Pattern Recognition (icIVPR)}.\hskip 1em
  plus 0.5em minus 0.4em\relax IEEE, 2019, pp. 324--329.

\bibitem{cote2019deep}
U.~C{\^o}t{\'e}-Allard, C.~L. Fall, A.~Drouin, A.~Campeau-Lecours, C.~Gosselin,
  K.~Glette, F.~Laviolette, and B.~Gosselin, ``Deep learning for
  electromyographic hand gesture signal classification using transfer
  learning,'' \emph{IEEE Transactions on Neural Systems and Rehabilitation
  Engineering}, vol.~27, no.~4, pp. 760--771, 2019.

\bibitem{cote2017transfer}
U.~Cote-Allard, C.~L. Fall, A.~Campeau-Lecours, C.~Gosselin, F.~Laviolette, and
  B.~Gosselin, ``Transfer learning for semg hand gestures recognition using
  convolutional neural networks,'' in \emph{2017 IEEE International Conference
  on Systems, Man, and Cybernetics (SMC)}.\hskip 1em plus 0.5em minus
  0.4em\relax IEEE, 2017, pp. 1663--1668.

\bibitem{lin2014human}
H.-I. Lin, M.-H. Hsu, and W.-K. Chen, ``Human hand gesture recognition using a
  convolution neural network,'' in \emph{2014 IEEE International Conference on
  Automation Science and Engineering (CASE)}.\hskip 1em plus 0.5em minus
  0.4em\relax IEEE, 2014, pp. 1038--1043.

\bibitem{liao2018hand}
B.~Liao, J.~Li, Z.~Ju, and G.~Ouyang, ``Hand gesture recognition with
  generalized hough transform and dc-cnn using realsense,'' in \emph{2018
  Eighth International Conference on Information Science and Technology
  (ICIST)}.\hskip 1em plus 0.5em minus 0.4em\relax IEEE, 2018, pp. 84--90.

\bibitem{yu2015multi}
F.~Yu and V.~Koltun, ``Multi-scale context aggregation by dilated
  convolutions,'' \emph{arXiv preprint arXiv:1511.07122}, 2015.

\bibitem{simonyan2014very}
K.~Simonyan and A.~Zisserman, ``Very deep convolutional networks for
  large-scale image recognition,'' \emph{arXiv preprint arXiv:1409.1556}, 2014.

\bibitem{schindler2016comparing}
A.~Schindler, T.~Lidy, and A.~Rauber, ``Comparing shallow versus deep neural
  network architectures for automatic music genre classification.'' in
  \emph{FMT}, 2016, pp. 17--21.

\bibitem{verma2019learnet}
M.~Verma, S.~K. Vipparthi, G.~Singh, and S.~Murala, ``Learnet: Dynamic imaging
  network for micro expression recognition,'' \emph{IEEE Transactions on Image
  Processing}, vol.~29, pp. 1618--1627, 2019.

\bibitem{verma2019hinet}
M.~Verma, S.~K. Vipparthi, and G.~Singh, ``Hinet: Hybrid inherited feature
  learning network for facial expression recognition,'' \emph{IEEE Letters of
  the Computer Society}, vol.~2, no.~4, pp. 36--39, 2019.

\bibitem{barczak2011new}
A.~Barczak, N.~Reyes, M.~Abastillas, A.~Piccio, and T.~Susnjak, ``A new 2d
  static hand gesture colour image dataset for asl gestures,'' 2011.

\bibitem{pugeault2011spelling}
N.~Pugeault and R.~Bowden, ``Spelling it out: Real-time asl fingerspelling
  recognition,'' in \emph{2011 IEEE International conference on computer vision
  workshops (ICCV workshops)}.\hskip 1em plus 0.5em minus 0.4em\relax IEEE,
  2011, pp. 1114--1119.

\bibitem{matilainen2016ouhands}
M.~Matilainen, P.~Sangi, J.~Holappa, and O.~Silv{\'e}n, ``Ouhands database for
  hand detection and pose recognition,'' in \emph{2016 Sixth International
  Conference on Image Processing Theory, Tools and Applications (IPTA)}.\hskip
  1em plus 0.5em minus 0.4em\relax IEEE, 2016, pp. 1--5.

\end{thebibliography}

\end{document}